\documentclass[runningheads]{llncs}
\usepackage[T1]{fontenc}
\usepackage{graphicx}

\usepackage{cite}
\usepackage{amsmath,amssymb,amsfonts}
\usepackage{textcomp}
\usepackage{xcolor}
\usepackage{subcaption}
\usepackage{makecell}
\usepackage{graphicx}
\usepackage{cite}
\usepackage{hyperref}
\usepackage{subcaption}
\usepackage{amsmath}
\usepackage{caption}
\usepackage{multirow}
\usepackage{booktabs}
\usepackage{enumerate}
\usepackage[margin=4cm]{geometry}

\begin{document}
\title{Advancing Parkinson's Disease Progression Prediction: Comparing Long Short-Term Memory Networks and Kolmogorov-Arnold Networks}

%
%
\author{Abhinav Roy\inst{1}\orcidID{0009-0000-6854-1274} \and
Bhavesh Gyanchandani\inst{2}\orcidID{0009-0008-9861-8568}\and
Aditya Oza\inst{3}\orcidID{0009-0003-6165-3611}\and
Abhishek Sharma\inst{4}\orcidID{}
}
%
%
\institute{IIIT Naya Raipur, India \\
\email{abhinav21102@iiitnr.edu.in, gyanchandani21102@iiitnr.edu.in,
aditya21102@iiitnr.edu.in,
abhishek@iiitnr.edu.in}\\}
\maketitle              
\begin{abstract}
Parkinson's Disease (PD) is a degenerative neurological disorder that impairs motor and non-motor functions, significantly reducing quality of life and increasing mortality risk. Early and accurate detection of PD progression is vital for effective management and improved patient outcomes. Current diagnostic methods, however, are often costly, time-consuming, and require specialized equipment and expertise. This work proposes an innovative approach to predicting PD progression using regression methods, Long Short-Term Memory (LSTM) networks, and Kolmogorov Arnold Networks (KAN). KAN, utilizing spline-parametrized univariate functions, allows for dynamic learning of activation patterns, unlike traditional linear models.

The Movement Disorder Society-Sponsored Revision of the Unified Parkinson's Disease Rating Scale (MDS-UPDRS) is a comprehensive tool for evaluating PD symptoms and is commonly used to measure disease progression. Additionally, protein or peptide abnormalities are linked to PD onset and progression. Identifying these associations can aid in predicting disease progression and understanding molecular changes.

Comparing multiple models, including LSTM and KAN, this study aims to identify the method that delivers the highest metrics. The analysis reveals that KAN, with its dynamic learning capabilities, outperforms other approaches in predicting PD progression. This research highlights the potential of AI and machine learning in healthcare, paving the way for advanced computational models to enhance clinical predictions and improve patient care and treatment strategies in PD management.
\end{abstract}
\section{Introduction}\label{sec1}

Parkinson's Disease (PD) is a progressive neurological disorder characterized by the gradual deterioration of motor and non-motor functions, profoundly impacting patients' quality of life. Accurate prediction of PD progression is vital for timely intervention and effective disease management. Time series forecasting in this context involves predicting future disease states based on historical clinical and physiological data, making it an indispensable tool for clinical decision-making.

Traditional approaches to forecasting PD progression have primarily relied on parametric models grounded in domain-specific knowledge, such as autoregressive (AR) models, exponential smoothing, and structural time series models. These methods, while effective in certain scenarios, often fall short in capturing the complex, nonlinear dynamics inherent in PD progression dataset. Additionally, traditional methods have integrated combination of regression techniques with deep learning models to enhance longitudinal forecasting accuracy. Techniques such as combining regression models with neural networks have been explored to leverage the strengths of both approaches, but they often struggle with scalability and interpretability.

The advent of machine learning (ML) has revolutionized time series forecasting by enabling the analysis of large datasets and uncovering intricate, nonlinear relationships that traditional models may overlook. In particular, ML models focused on PD classification have demonstrated remarkable success due to their flexibility and adaptability. Subsequently, more sophisticated architectures, including Long Short-Term Memory (LSTM) networks and Gated Recurrent Units (GRUs), were developed to address the limitations of traditional recurrent networks, such as the vanishing gradient problem. Additionally, Convolutional Neural Networks (CNNs), known for their proficiency in spatial pattern recognition, have been employed in conjunction with LSTMs to efficiently process both spatial and temporal dependencies, thereby enhancing forecasting accuracy.

Despite their advancements, many DL models, face challenges related to scalability and interpretability. The number of parameters in these networks often grows non-linearly with the addition of layers, complicating their implementation and understanding. Otherwise, they just classify the type of Parkinson instead of specifying how much progression happened. 

A new neural network design called Kolmogorov-Arnold Networks (KANs) has emerged as a potential game changer. KANs, which are inspired by the Kolmogorov-Arnold representation theorem, use spline-based univariate functions that are constructed as learnable activation functions instead of standard linear weights. This novel design improves the network's accuracy and interpretability, allowing it to obtain equivalent or greater outcomes with smaller network sizes across a variety of applications.

The present study looks at the use of KANs to forecast longitudinal datasets of Parkinson's disease progression, an area that has not been well studied in the literature. We intend to assess the usefulness of KANs in real-world clinical scenarios, examining their efficiency in terms of the number of trainable parameters and exploring how their additional degrees of freedom may affect forecasting ability. Using real-world clinical data, we evaluate the performance of KANs with classic regression models and advanced DL architectures such as LSTMs and other regression approaches to determine the most effective way to predict Parkinson's disease development. This study aims to verify KANs as a versatile tool in advanced neural network design for clinical time series forecasting, although more research is needed to maximize their use in broader applications.

Major contributions of this research are given below :
\begin{itemize}
    \item The study introduces the use of Kalmogorov-Arnold Networks (KAN) to analyze AMP MDS UPDRS datasets, demonstrating exceptional performance in capturing the longitudinal progression of Parkinson’s disease and marks the first use of KAN for predicting disease progression in this specific context, showcasing its ability to effectively model and interpret complex temporal patterns in clinical data.

    \item This research presents a detailed comparative analysis of various methodologies, including traditional regression techniques, KAN, and state-of-the-art deep learning approaches. By systematically evaluating these methods, the study highlights the strengths and limitations of each, ultimately identifying the most effective approach for predicting Parkinson’s disease progression based on UPDRS scores. 
    
    \item The study innovatively develops and integrates lag variables into predictive models for Parkinson’s disease progression, linking historical clinical features with future UPDRS scores. This methodological advancement captures the temporal dynamics of disease progression, offering a more nuanced and effective framework for forecasting future clinical states, thus improving predictive accuracy and robustness.

\end{itemize}
The remainder contents of this work are framed as follows: Section II contains the literature review, exploring research gaps and challenges. Section III discusses different methodologies used, including dataset description and proposed methods. Section IV presents a comprehensive discussion about results, while Section V concludes the work .

\section{Related Works}\label{sec2}
Numerous studies have explored the potential of ML techniques in diagnosing PD using various data types, particularly speech signals and medical imaging. These approaches offer non-invasive, accurate diagnostic tools that can significantly enhance the early detection and management of PD. Several studies have focused on the analysis of speech signals for PD diagnosis. Celik and Omurca compared various ML classification methods, including Logistic Regression, Support Vector Machines (SVM), and Random Forests, to predict PD from speech data~\cite{celik2019improving}. By expanding feature sets using Principal Component Analysis (PCA) and Information Gain (IG), they achieved improved classification accuracy. Similarly, Cantürk and Karabiber developed a comprehensive ML system that utilized multiple feature selection algorithms and classifiers. They emphasized the medical significance of selected features, achieving high accuracy in distinguishing PD patients from healthy individuals~\cite{canturk2016machine}.

Sakar et al. investigated different types of voice samples, such as sustained vowels, words, and sentences, and found that sustained vowels were particularly effective in predicting PD. They also noted that using central tendency and dispersion metrics improved the generalization of their predictive model~\cite{sakar2013collection}. Tsanas et al. tested novel dysphonia measures and identified specific subsets that significantly enhanced classification accuracy, reaching nearly 99\% accuracy with only ten features \cite{tsanas2012novel}. Frid et al. demonstrated that PD diagnosis could be achieved directly from analog speech signals, differentiating between various stages of the disease using a combination of signal processing and ML techniques~\cite{frid2014computational}. Sztahó et al. applied linear regression models to acoustic features from speech samples of varying lengths, participating in the INTERSPEECH 2015 challenge to perform automatic PD classification~\cite{sztaho2015estimating}.

In addition to speech signal analysis, medical imaging has also been utilized for PD diagnosis. Wingate et al. proposed a deep learning approach using convolutional and recurrent neural networks to analyze medical images, such as MRI and dopamine transporter scans. Their unified framework demonstrated effective prediction across different medical environments, illustrating the potential of deep learning in PD diagnosis~\cite{wingate2020unified}.

The Movement Disorder Society-sponsored revision of the Unified Parkinson's Disease Rating Scale (MDS-UPDRS) provides a comprehensive assessment of PD. Goetz et al. presented the MDS-UPDRS, which includes evaluations of non-motor and motor experiences, motor examination, and complications. The scale showed high internal consistency and strong correlations with the original UPDRS, supporting its validity in rating PD~\cite{goetz_movement_2008}.

The use of metabolomics in PD diagnosis has also been explored. Zhang et al. employed an interpretable neural network framework to predict PD from blood plasma metabolomics data. Their approach identified significant biomarkers, including exogenous substances, that predate clinical diagnosis and contribute to early disease prediction, highlighting the potential of metabolomics in PD research \cite{zhang2023interpretable}.

Voice disorders, collectively termed hypokinetic dysarthria, are characteristic of PD. Suppa et al. investigated voice changes in PD patients at different stages of the disease. Using SVM classifiers to analyze voice samples, they found that voice abnormalities were present in early-stage PD and progressively worsened. L-Dopa therapy improved but did not completely restore voice quality. Machine learning provided high accuracy in tracking disease severity and evaluating the therapeutic effects, suggesting that voice analysis could serve as a potential biomarker for PD progression~\cite{suppa2022voice}.

Despite the promising advances in PD diagnosis using ML and various data types, a significant research gap exists. Most studies have primarily focused on the classification and early diagnosis of PD rather than on tracking its progression. While accurate diagnosis is essential, understanding and monitoring disease progression is equally crucial for effective management and treatment planning. The progression of PD varies significantly among individuals, and continuous monitoring can provide insights into disease dynamics and the effectiveness of therapeutic interventions. However, there is a paucity of research dedicated to developing models and systems specifically designed to track PD progression over time.

\section{Methodology}\label{sec6}
\begin{figure}[!htbp]
    \centering
    \includegraphics[width=0.85\textwidth]{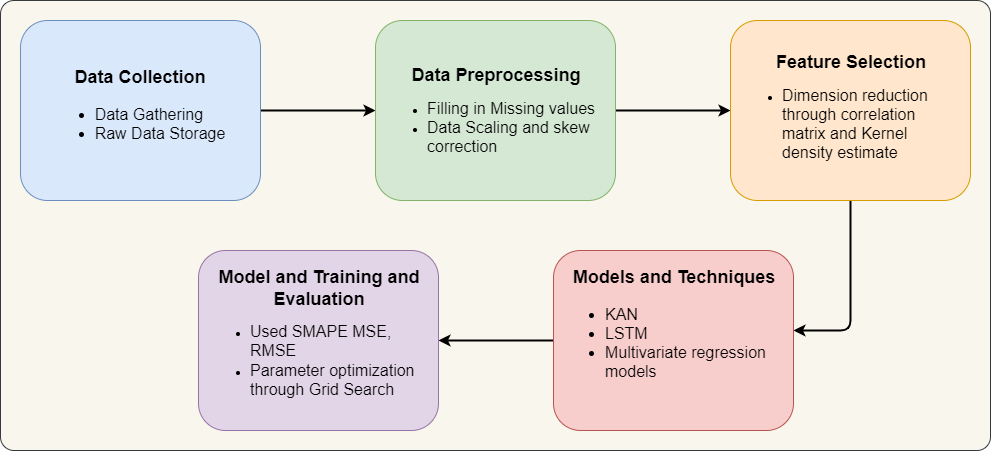}
    \caption{Proposed Model with KAN and LSTM}
    \label{fig:PMLSTM}
\end{figure}

To address the objective of parkinson progression  based on MDS UPDRS score, the study provides a comprehensive methodology consisting of several key steps. First, dataset description is there, then work on our model starting with data preprocessing, extraction of features from a given dataset, the detailed working of different types of models used and finally, a description of model parameters . Figure~\ref{fig:PMLSTM} demonstrates our plan of action

\subsection{Dataset description}
The dataset used in this research focuses on predicting Parkinson's disease progression. It contains data from 248 patients who underwent regular Cerebrospinal Fluid Mass Spectrometry (CSF-MS) tests and Movement Disorder Society Unified Parkinson's Disease Rating Scale (MDS-UPDRS) scores at multiple time points (0, 6, 12, and 24 months). This time-series data is of longitudinal nature and allows us to use Deep learning techniques like LSTM or KAN to predict future MDS-UPDRS scores.~\cite{goetz2010movement}

The core of the dataset is protein abundance information derived from CSF samples analyzed using mass spectrometry. Table\ref{tab:data_sources} demonstrates us about more information about the dataset. Each patient contributed multiple samples over several years, with corresponding MDS-UPDRS assessments performed to gauge Parkinson's disease severity at each time point.
 The MDS-UPDRS score which represents the motor symptoms and clinometric characteristics of Parkinson's disease (PD) on a scale, is one of the dataset's features ranges from 0 to 272, where 0 denotes normalcy and 272 major motor and non-motor deterioration. While the complete set of proteins linked to PD is still under investigation, those demonstrating potential for predicting disease progression warrant further study.

\begin{table}[htbp]
    \centering
    \caption{Data Sources and Descriptions}
    \label{tab:data_sources}
    \begin{tabular}{|p{2.3cm}|p{2.5cm}|p{2cm}|p{4cm}|p{3cm}|}
        \hline
        \textbf{Data Source} & \textbf{Description} & \textbf{Time Point} & \textbf{Key Information} & \textbf{Connection to Patients} \\
        \hline
        train\_peptides csv file & Mass spectrometry data & Visit Month & Peptide sequences, abundance, and associated protein (UniProt ID) & Solid line (data collected from patients) \\
        \hline
        train\_proteins csv file & Aggregated protein expression data & Visit Month & Normalized protein expression (NPX) for each UniProt ID & Solid line (data collected from patients) \\
        \hline
        train\_clinical\_-data csv file & Clinical data & Visit Month & UPDRS scores (1-4) assessing PD severity, medication status & Solid line (clinical assessments of patients) \\
        \hline
        supplemental\_-clinical\_data csv file & Supplemental clinical data & Not applicable & Additional clinical records for context & - \\
        \hline
    \end{tabular}
\end{table}

\subsection{Data preprocessing }\label{sec7}
The data was explored using Python Jupyter notebook, a report was created to understand the dataset more deeply with the help of Pandas and Numpy Library. The dataset consisted of 3 types, categorical values for UniProt, Peptide, visit id and upd23b clinical state on medication, integer values for patient\_id and visit month and for rest of features it was float value.
In the report, we find out that about 20.58\% data is skewed , and 8.9\% data is missing, with UPDRS-IV features having the most missing values. We also standardize data to have a consistent scale. Also we apply one hot encoding to convert upd23b into numerical data type for easier analysis.
  \begin{enumerate}[(1)]
  \item \textbf{Skewness Removal:} We apply logarithmic transformation, which helps in mitigating 13.92\% of skewness, with Box-Cox 6\% skewness mitigated and rest through square root transformations.
  \item \textbf{Standardization:} We standardize the features by centering them around zero average and a standard deviation of one.
  \item \textbf{Handling Missing Values:} We find out about missing values based on their type, like Missing completely at random (MCAR), which gets imputed by mean imputation. For  Missing at random (MAR), and Missing not at random (MNAR) gets imputed through soft impute. through various imputation techniques, including Singular Value Decomposition (SVD) for MAR and MNAR cases.
  \item \textbf{One-hot encoding for upd23b:} It  has three categories: On, Off, and NaN (No). NaN indicates that Levodopa was not administered, On indicates a positive effect after administration, and Off indicates an ineffective result even after administration. To make analysis easy, this was transformed into three separate columns using One-hot encoding.
\end{enumerate}

\subsection{Feature selection}\
Feature selection is crucial in identifying significant predictors for the UPDRS scores prediction which tells us about Parkinson progression.
\subsubsection{Correlation Matrix }
\begin{figure}[!htbp]
    \centering
    \includegraphics[width=0.7\textwidth]{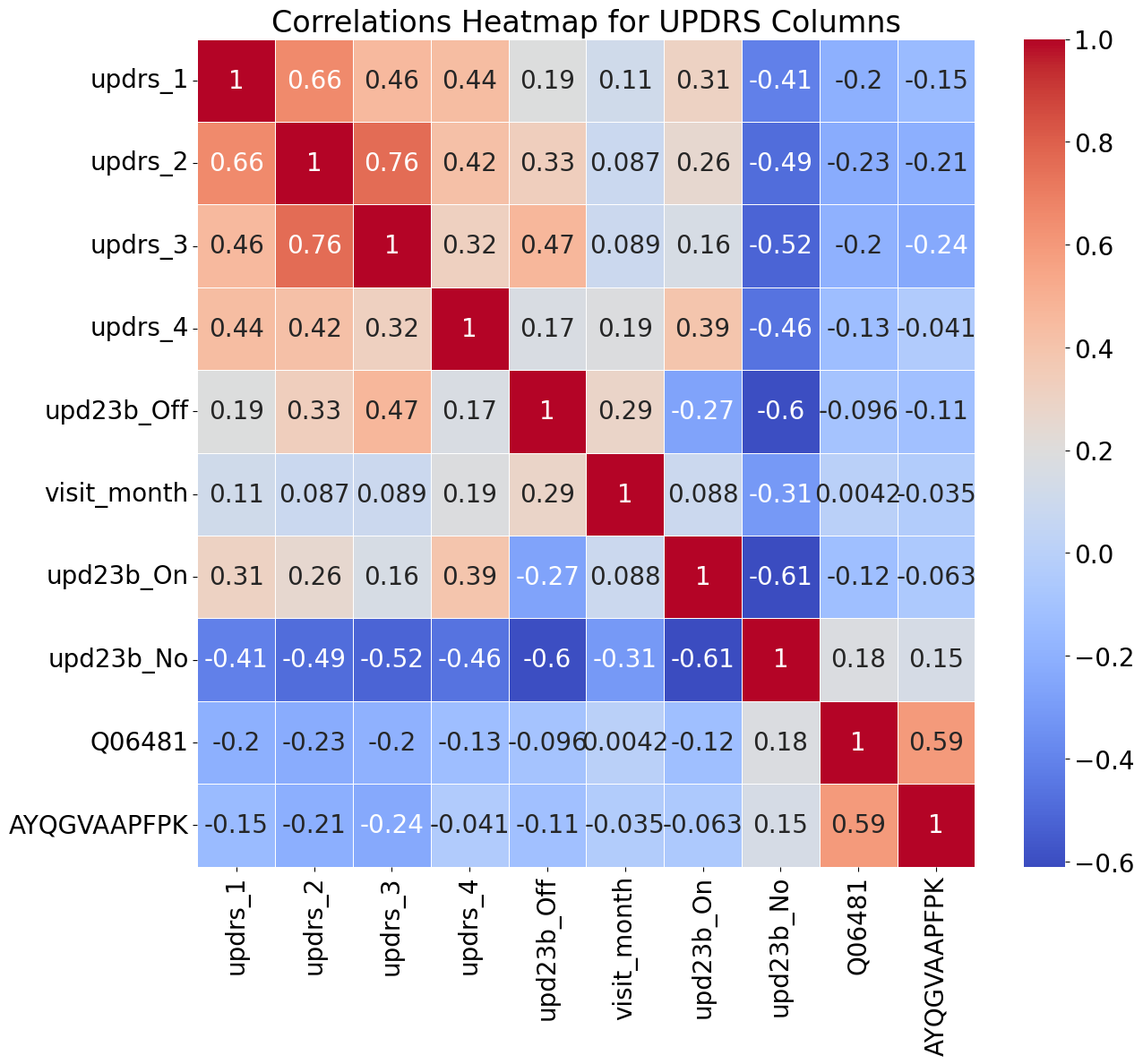}
    \caption{Correlation matrix for merged dataset}
    \label{fig:corr}
\end{figure}
From Figure \ref{fig:corr}, we observe the correlations among various UPDRS columns and other variables. The correlation matrix provides insights into the relationships between different UPDRS scores and other variables. Strong correlations may indicate potential predictors that could be important for our predictive model. For instance, UPDRS\_1 and UPDRS\_2 exhibit a strong positive correlation of 0.66, suggesting a close relationship between these two scores. Like that, we can say UPDRS\_2 and UPDRS\_3 are highly correlated. Overall we can conclude from the correlation matrix that is the use of previous UPDRS scores and visit\_month as the most informative features to predict future UPDRS scores.

\subsubsection{Kernel Density Estimate}

\begin{figure}[!htbp]
    \centering
    \includegraphics[width=1\textwidth]{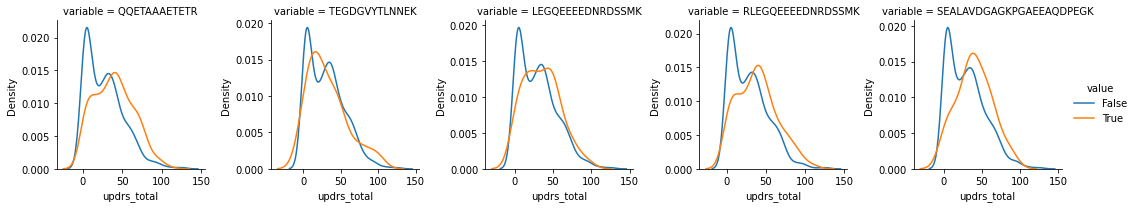}
    \caption{Density plots for five different variables peptides against total UPDRS }
    \label{fig:KDE}
\end{figure}
We employed a statistical analysis to evaluate the importance of peptides in relation to UPDRS scores. To visualize the distributions, we utilized a Kernel Density Estimate as shown in Figure \ref{fig:KDE}. Figure \ref{fig:KDE} shows clearly important differences in distribution of UPDRS, depending on the presence of some peptides. This means that we need a predictive method that can treat the simple presence of a peptide (ignoring the concentration) as important for the prediction.
 This observation suggests that even the simple presence of a peptide, regardless of concentration, may be important for predicting UPDRS scores.

Based on these findings, our feature selection process will focus on including peptide presence indicators as potential predictors in our predictive model with previous UPDRS values and visit\_month. This approach acknowledges the observed importance of peptide occurrence in distinguishing UPDRS scores, thereby enhancing the predictive power of our model.

\subsection{Model training}

\begin{table}[htbp]
\centering
\begin{minipage}{0.45\textwidth}
\caption{Model Summary for LSTM}
\centering
\begin{tabular}{|p{2cm}|p{1.7cm}|p{1.9cm}|}
\hline
\textbf{Type} & \textbf{Hidden Size} & \textbf{Parameters} \\
\hline
LSTM & 64.0  & 246,272 \\
\hline
Attention & 128.0 &  16,641 \\
\hline
FC &  128.0 &  8,256 \\
\hline
BatchNorm &  32.0 &  64 \\
\hline
FC & 32.0 &  528 \\
\hline
BatchNorm &  16.0 & 32 \\
\hline
FC & 16.0 &  68 \\
\hline
\textbf{Total} & Output-4 & \textbf{271,861} \\
\hline
\end{tabular}

\label{tab:model_layers_lstm_attention}
\end{minipage}%
\hspace{0.05\textwidth} 
\begin{minipage}{0.45\textwidth}
\caption{Model Summary for KAN}
\centering
\begin{tabular}{|p{2cm}|p{1.7cm}|p{1.9cm}|}
\hline
\textbf{Type} & \textbf{In Features} & \textbf{Parameters} \\
\hline
Input Layer & 27 & 20,655 \\
\hline
KAN Layer & 45 & 69,615 \\
\hline
KAN Layer & 91 &  283,101 \\
\hline
Output & 183 &736 \\
\hline
\textbf{Total} &Output-4 & \textbf{374,107} \\
\hline
\end{tabular}
\label{tab:model_layers_kan}
\end{minipage}
\end{table}

The primary objective of training the LSTM model was to learn representations of sequential data that could be used for accurate prediction future UPDRS scores by using previous UPDRS scores and peptides value. The model was configured with a hidden state dimension of 64, which was crucial for capturing complex patterns in the data. A dropout rate of 0.2 was applied to the LSTM layer to prevent overfitting by introducing regularization and promoting generalization to unseen data.

To optimize the model's performance, the Mean Squared Error (MSE) loss function was employed, as it is a standard choice for regression tasks where the goal is to minimize the difference between predicted and actual values. The Adam optimizer was used with a learning rate of 0.001, a commonly effective setting for balancing convergence speed and stability. Weight decay was included to further prevent overfitting by penalizing large weights.

The training process was managed over a maximum of 500 epochs, with early stopping implemented to halt training when the validation loss did not improve for 50 consecutive epochs. This approach prevented the model from overfitting to the training data and ensured that training was stopped at the point of best generalization.

Meanwhile, the goal of training the KAN model was to harness the theoretical advantages of Kolmogorov-Arnold networks for function approximation tasks. The model’s architecture utilized a grid size of 10 and a spline order of 3 to effectively capture intricate relationships within the data through a combination of approximation and spline functions. A dropout rate of 0.2 was employed to regularize the model and prevent overfitting.

For the training process, the Mean Squared Error (MSE) loss function was again used, as it is well-suited for regression tasks where the objective is to minimize the average squared differences between predicted and actual values. The Adam optimizer, configured with a learning rate of 0.0005 and weight decay of 1e-5, was selected for its robustness in optimizing network parameters and controlling overfitting.

Similar to the training of LSTM the training was conducted over a maximum of 500 epochs, with early stopping implemented to cease training if there was no improvement in validation loss for 50 epochs. This strategy aimed to find the optimal balance between model complexity and performance, ensuring that the model did not overfit while achieving the best possible validation results.

\begin{figure}[tb]
  \centering
  \begin{subfigure}{0.48\linewidth}
    \includegraphics[width=1\textwidth]{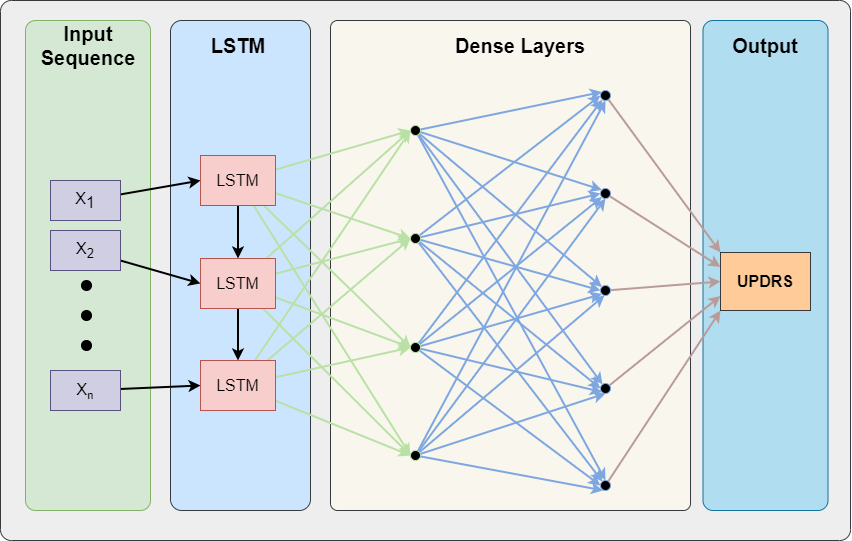}
    \caption{Structure of LSTM}
    \label{fig:lstms}
  \end{subfigure}
  \hfill
  \begin{subfigure}{0.48\linewidth}
    \includegraphics[width=1\textwidth]{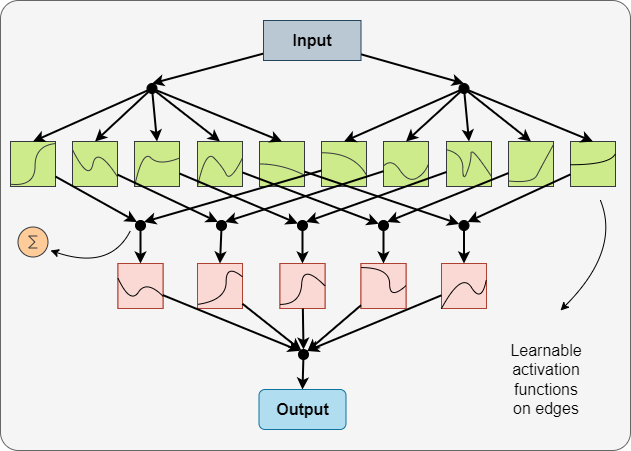}
    \caption{Structure of KAN}
    \label{fig:kans}
  \end{subfigure}
  \caption{Structures of our models}
  \label{fig:structures}
\end{figure}

In summary, both the LSTM and KAN models were trained with methodologies tailored to their specific strengths for predicting future 4 UPDRS scores. The LSTM model utilized its recurrent architecture to capture temporal dependencies in sequential data, with dropout and early stopping strategies employed to balance learning and generalization. The KAN model leveraged its theoretical foundation in function approximation by the Kolmogorov-Arnold representation theorem to achieve accurate predictions of future 4 UPDRS scores. The structures of our model can be seen in Figure~\ref{fig:structures}. Both models used MSE loss and the Adam optimizer, with careful consideration of training epochs and early stopping criteria to ensure effective model training and performance.

\section{Results and discussion}\label{sec8}

\subsection{Evaluation metrics}
To evaluate the supremacy of the proposed framework, we have utilized the following statistical performance evaluation metrics:
\textbf{Symmetric Mean Absolute Percentage Error (SMAPE)}\cite{Thieu_PerMetrics_A_Framework_2024}\cite{van2023mealpy}:This metric is used to evaluate how accurate forecasting techniques are. The average error is expressed as a percentage of the difference between the actual and forecast values' absolute values.

\begin{equation}
\text{SMAPE}(y, \hat{y}) = \frac{100\%}{N} \sum_{i=0}^{N - 1} \frac{ 2*|y_i - \hat{y}_i|}{|y| + |\hat{y}|}
\end{equation}
where: \( N \) = number of samples,
\( y_i \) = actual value at instance \( i \),
\( \hat{y}_i \) = predicted value at instance \( i \).

\textbf{Mean Square Error(MSE)}\cite{mediumArticle}: This metric is used to evaluate the accuracy of forecasting techniques by measuring the average squared difference between the actual and forecast values.

\begin{equation}
\text{MSE}(y, \hat{y}) = \frac{1}{N} \sum_{i=0}^{N - 1} (y_i - \hat{y}_i)^2
\end{equation}
where: \( N \) = number of samples,  
\( y_i \) = actual value at instance \( i \),  
\( \hat{y}_i \) = predicted value at instance \( i \).

\textbf{Root Mean Squared Error (RMSE)}: RMSE, akin to MSE \cite{mediumArticle}, incorporates the square root to provide a measure of average error magnitude in the same units as the dependent variable.
\begin{equation}
\text{RMSE} = \sqrt{\frac{1}{n} \sum_{i=1}^{n} \left( y_i - \hat{y}_i \right)^2}
\end{equation}
where:
\( n \) = number of samples,
\( y_i \) = actual value at instance \( i \),
\( \hat{y}_i \) = predicted value at instance \( i \).

\subsection{Evaluation of model}
Now we evaluate our research through evaluation metrics mentioned .
\begin{figure}[!htbp]
  \centering
  \begin{minipage}[t]{0.49\linewidth}
    \centering
    \includegraphics[width=\linewidth]{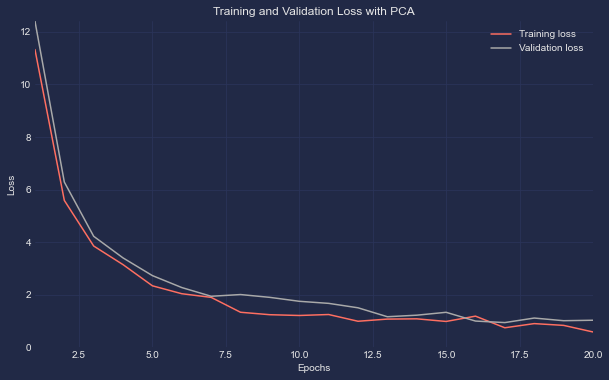}
    \caption{Plot of train and validation loss of LSTM}
    \label{LSTM_t}
  \end{minipage}
  \hfill
  \begin{minipage}[t]{0.49\linewidth}
    \centering
    \includegraphics[width=\linewidth]{Figures/KAN.png}
    \caption{Plot of train and validation loss of KAN model}
    \label{KAN_t}
  \end{minipage}
\end{figure}

 Firstly, we compare the training and validation graph. As shown in Figure~\ref{KAN_t}, training and validation loss are both lower and the curve is smoother when KAN is used compared to the model which uses LSTM (Figure~\ref{LSTM_t}). This suggests that KAN deals with the features more efficiently than LSTM, leading to a better fit and potentially improved generalizability.

\begin{table}[htbp]
    \centering
    \begin{minipage}{0.45\textwidth}
        \caption{Performance Metrics of Model with LSTM on validation data}
        \centering
        \begin{tabular}{|l|c|c|c|}
            \hline
            \textbf{Target} & \textbf{SMAPE}  & \textbf{MSE} & \textbf{RMSE}\\
            \hline
            UPDRS\_1 & 61.18 &19.58 & 4.424\\
            \hline
            UPDRS\_2 & 84.78 &26.14 & 5.112\\
            \hline
            UPDRS\_3 & 75.68 & 112.19 & 10.592\\
            \hline
            UPDRS\_4 & 157.43 & 9.47 & 3.077\\
            \hline
            Average & 94.77 & 41.85 & 5.801\\
            \hline
        \end{tabular}
        \label{lstm_metrics}
    \end{minipage}
    \hspace{0.05\textwidth} 
     \begin{minipage}{0.45\textwidth}
        \caption{Performance Metrics of Model with KAN on validation data}
        \centering
        \begin{tabular}{|l|c|c|c|}
            \hline
        \textbf{Target} & \textbf{SMAPE}  & \textbf{MSE} & \textbf{RMSE}\\ 
            \hline
            UPDRS\_1 & 50.64 &14.48 & 3.805 \\
            \hline
            UPDRS\_2 & 78.75 & 22.88 & 4.783\\
            \hline
            UPDRS\_3 & 69.49 &84.07 & 9.168\\
            \hline
            UPDRS\_4 & 172.44 &7.34 & 2.709\\
            \hline
            Average & 92.83 & 32.19 & 5.116\\
            \hline
        \end{tabular}
        \label{KAN_metrics}
    \end{minipage}
\end{table}

Table~\ref{lstm_metrics} displays the results of the LSTM model, which include an average RMSE of 5.801, an average MSE of 41.85, and an average SMAPE of 94.77. For the target UPDRS\_1, the LSTM model yielded the best results. Table~\ref{KAN_metrics} presents the results of the KAN model. It reveals an average RMSE of 5.116, an average MSE of 32.19, and an average SMAPE of 92.83. It also performed best for the UPDRS\_1 target variable.

\begin{table}[htbp]
    \centering
    \caption{Benchmarking model final results for Parkinson progression prediction using UPDRS}
    \label{tab:benchmark}
    \begin{tabular}{|p{2cm}|p{2.3cm}|p{2cm}|p{2cm}|p{3cm}|}
        \hline
        \textbf{Model} & \textbf{Avg SMAPE} & \textbf{Avg MSE} & \textbf{Avg RMSE} & \textbf{Training time(s)}\\
        \hline
        LSTM & 94.76 & 41.85 & 5.801 & 501 \\
        \hline
        KAN & 92.83 & 32.19 & 5.116 & 1208\\
        \hline
    \end{tabular}
\end{table}

Overall, Table~\ref{tab:benchmark} presents the benchmark following an assessment and comparison of the two models' performances. When RMSE, MSE, and SMAPE were taken into account, KAN was the model that performed the best, yielding average figures of 5.116, 32.19, and 92.83, respectively. The LSTM, with RMSE, MSE, and SMAPE of 5.801, 41.85, and 94.76, produced outcomes that were comparable to those of the KAN.
In terms of training time computation, LSTM takes 501 s, whereas KAN takes 1208 s.
Overall, we can say that KAN was the model that did best for this task based on the criteria provided by the challenge, which include SMAPE, MSE, and RMSE. However, LSTM would be the best model for this task when processing time is taken into account.

\begin{figure}[!htbp]
  \centering
  \begin{subfigure}{0.48\linewidth}
    \includegraphics[width=\linewidth]{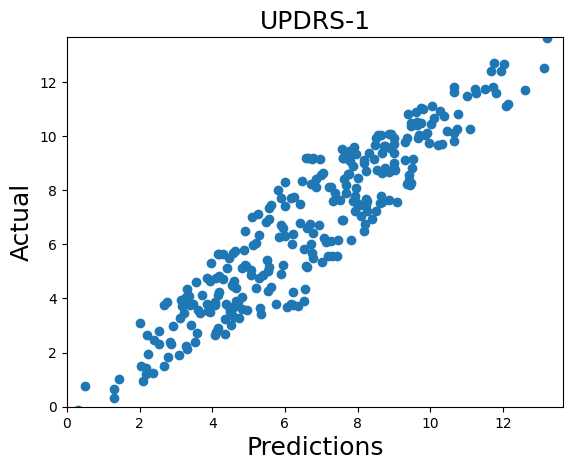}
    \caption{UPDRS 1 SMAPE }
    \label{fig:updrs1_smape}
  \end{subfigure}
  \hfill
  \begin{subfigure}{0.48\linewidth}
    \includegraphics[width=\linewidth]{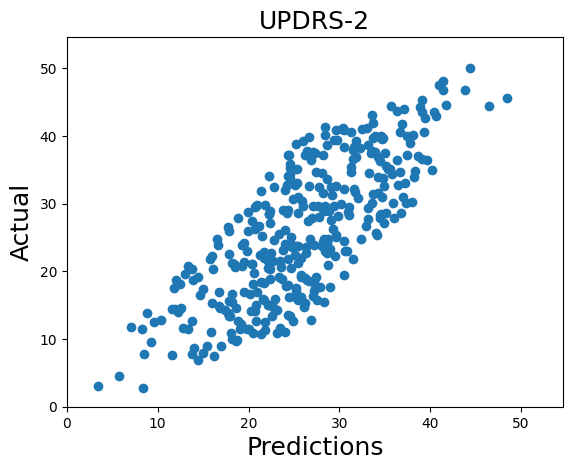}
    \caption{UPDRS 2 SMAPE }
    \label{fig:updrs2_smape}
  \end{subfigure}
  
  \begin{subfigure}{0.48\linewidth}
    \includegraphics[width=\linewidth]{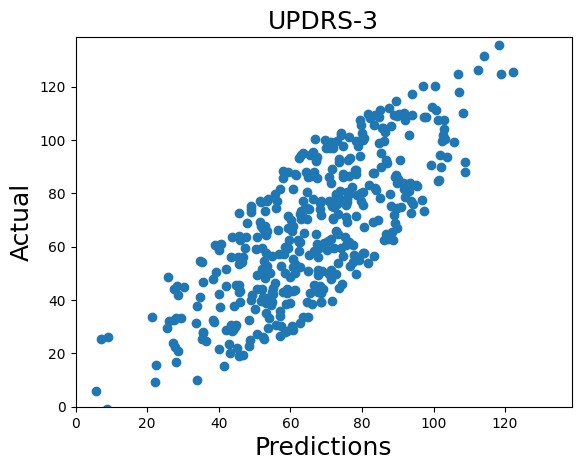}
    \caption{ UPDRS 3 SMAPE }
    \label{fig:updrs3_smape}
  \end{subfigure}
  \hfill
  \begin{subfigure}{0.48\linewidth}
    \includegraphics[width=\linewidth]{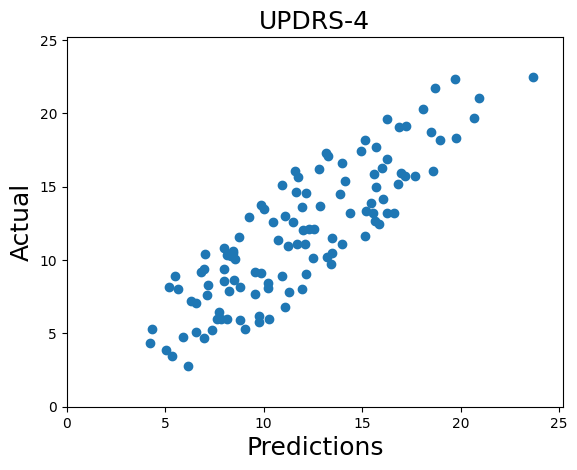}
    \caption{UPDRS 4 SMAPE }
    \label{fig:updrs4_smape}
  \end{subfigure}

  \caption{SMAPE plot with our KAN  model}
  \label{fig:LSTM-PCA model}
\end{figure}

Also we try to show why our SMAPE for KAN are coming good through  Figure~\ref{fig:updrs1_smape} , Figure~\ref{fig:updrs2_smape},  Figure~\ref{fig:updrs3_smape} and  Figure~\ref{fig:updrs4_smape}  correspond to four different measures of the "Unified Parkinson's Disease Rating Scale (UPDRS)" "Symmetric Mean Absolute Percentage Error (SMAPE)". UPDRS is a clinical tool used to assess the severity of Parkinson's disease. The four measures shown here are UPDRS parts 1, 2, 3, and 4, which assess different aspects of the disease.~\cite{vladik2014estimate}

SMAPE is a way of measuring the accuracy of prediction models. A perfect prediction would have a SMAPE of 0\%. The closer a SMAPE value is to zero, the better the model's performance. SMAPE values range from 0\% to 100\%. A lower SMAPE value indicates better predictive accuracy, with:
\begin{itemize}
   
 \item 0\% SMAPE: This indicates perfect prediction, with no errors between the forecasted and actual values.
\item Low SMAPE (typically below 10\%): Generally considered excellent forecasting accuracy.
\item Moderate SMAPE (between 10\% and 20\%): Indicates good to reasonable forecasting performance.
\item High SMAPE (above 20\%): Reflects poor model performance or significant differences between forecasted and actual values.
 \item What counts as a "good" SMAPE can depend on the specific context and the typical error margins in the relevant industry or domain\cite{mediumArticle1}.

\end{itemize}
   The y-axis in the provided plot shows the values that the KAN model predicted, and the x-axis shows the actual UPDRS values. The data points should ideally fall at a 45-degree angle along a straight diagonal line. This would suggest that the model's predictions and the actual UPDRS values match exactly.

While the model predictions aren't a perfect match for every data point (shown by the scatter), there's a positive trend. This indicates the model captures the overall progression of the UPDRS scores. It's learning the general relationship between features and UPDRS values, even if it can't pinpoint every single data point exactly. This is a promising sign for fine refinement and potentially strong performance of our model

\subsection{Comparative Analysis}

\begin{table}[h!]
    \centering
    \caption{Performance Metrics Comparison Across Studies}
    \begin{tabular}{|p{3cm}|p{3cm}|p{3.5cm}|p{1.6cm}|p{1.6cm}|p{1.6cm}|}
        \hline
        \textbf{Study} & \textbf{Model Type} & \textbf{Configuration} & \textbf{Avg SMAPE} & \textbf{Avg MSE} & \textbf{Avg RMSE} \\
        \hline
        \textbf{Our Study} & \textbf{LSTM} & - & \textbf{94.76} & \textbf{41.85} & \textbf{5.801} \\
        \cline{2-6}
                  & \textbf{KAN} & \textbf{-} & \textbf{92.83} & \textbf{32.19} & \textbf{5.116} \\
        \hline
        Ahmed Siraj~\cite{ahmed2020prediction} & Vanilla RNN & With Dense Connections & N/A & N/A & 6.01 \\
        \cline{2-6}
                    & LSTM & With Dense Connections & N/A & N/A & 6.22 \\
        \hline
        Oscar Andres Mancha Mendoza~\cite{guadalajarabenchmarking} & Random Forest & Average & 89.44 & 83.15 & N/A \\
        \hline
    \end{tabular}
\end{table}

Our KAN model demonstrates superior performance compared to other models, achieving the lowest average RMSE (5.116) and MSE (32.19). In comparison, Ahmed Siraj's best model, the Vanilla RNN with dense connections, has a higher average RMSE (6.01), and the Random Forest model in Oscar Andres Mancha Mendoza's study exhibits higher average MSE (83.15) and less consistent SMAPE (89.44).

The KAN model's lower RMSE and MSE indicate better predictive accuracy, making it a more reliable choice for time series forecasting tasks. Despite the longer training time (1208 seconds), the accuracy benefits of the KAN model justify its use. In contrast, while the LSTM in our study has a faster training time (501 seconds), its slightly higher RMSE (5.801) makes KAN the preferable option when accuracy is paramount.

\section{Conclusion}\label{sec8}
This study has investigated the application of machine learning techniques to predict the progression of Parkinson's disease (PD) using the MDS-UPDRS score. By leveraging longitudinal data from 248 patients, including protein abundance information derived from CSF samples and corresponding MDS-UPDRS assessments, we developed models to forecast future disease severity.

Our methodology included comprehensive data preprocessing, feature selection, and model training using Long Short-Term Memory (LSTM) networks and Kolmogorov-Arnold Networks (KAN). The results demonstrated the effectiveness of these models in capturing temporal dependencies and complex relationships within the data. Specifically, the LSTM model's recurrent architecture and the KAN model's function approximation capabilities proved valuable in making accurate predictions.

Among the models evaluated, the KAN model emerged as the best-performing approach. Its ability to dynamically learn activation patterns through spline-parametrized univariate functions allowed for superior accuracy in predicting Parkinson's disease progression compared to traditional regression techniques and LSTM models. This superiority highlights KAN's potential to model complex, non-linear relationships inherent in PD progression data.

In summary, our findings underscore the potential of machine learning in advancing PD research and support the development of predictive models that can aid in early diagnosis and monitoring of disease progression. The demonstrated efficacy of the KAN model, in particular, marks a significant step towards enhancing predictive capabilities and ultimately contributing to better patient outcomes.

%
%
%
%
\bibliographystyle{splncs04}
\bibliography{main}

\begin{thebibliography}{10}
\providecommand{\url}[1]{\texttt{#1}}
\providecommand{\urlprefix}{URL }
\providecommand{\doi}[1]{https://doi.org/#1}

\bibitem{celik2019improving}
Celik, E., Omurca, S.I.: Improving parkinson's disease diagnosis with machine learning methods. In: 2019 Scientific Meeting on Electrical-Electronics \& Biomedical Engineering and Computer Science (EBBT). pp.~1--4. Ieee (2019)

\bibitem{canturk2016machine}
Cant{\"u}rk, {\.I}., Karabiber, F.: A machine learning system for the diagnosis of parkinson’s disease from speech signals and its application to multiple speech signal types. Arabian Journal for Science and Engineering  \textbf{41},  5049--5059 (2016)

\bibitem{sakar2013collection}
Sakar, B.E., Isenkul, M.E., Sakar, C.O., Sertbas, A., Gurgen, F., Delil, S., Apaydin, H., Kursun, O.: Collection and analysis of a parkinson speech dataset with multiple types of sound recordings. IEEE Journal of Biomedical and Health Informatics  \textbf{17}(4),  828--834 (2013)

\bibitem{tsanas2012novel}
Tsanas, A., Little, M.A., McSharry, P.E., Spielman, J., Ramig, L.O.: Novel speech signal processing algorithms for high-accuracy classification of parkinson's disease. IEEE transactions on biomedical engineering  \textbf{59}(5),  1264--1271 (2012)

\bibitem{frid2014computational}
Frid, A., Safra, E.J., Hazan, H., Lokey, L.L., Hilu, D., Manevitz, L., Ramig, L.O., Sapir, S.: Computational diagnosis of parkinson's disease directly from natural speech using machine learning techniques. In: 2014 IEEE international conference on software science, technology and engineering. pp. 50--53. IEEE (2014)

\bibitem{sztaho2015estimating}
Sztah{\'o}, D., Kiss, G., Vicsi, K.: Estimating the severity of parkinson's disease from speech using linear regression and database partitioning. In: INTERSPEECH. pp. 498--502 (2015)

\bibitem{wingate2020unified}
Wingate, J., Kollia, I., Bidaut, L., Kollias, S.: Unified deep learning approach for prediction of parkinson's disease. IET Image Processing  \textbf{14}(10),  1980--1989 (2020)

\bibitem{goetz_movement_2008}
Goetz, C.G., et~al.: Movement disorder society-sponsored revision of the unified parkinson's disease rating scale (mds-updrs): Scale presentation and clinimetric testing results. Movement Disorders  \textbf{23},  2129--2170 (2008)

\bibitem{zhang2023interpretable}
Zhang, J.D., Xue, C., Kolachalama, V.B., Donald, W.A.: Interpretable machine learning on metabolomics data reveals biomarkers for parkinson’s disease. ACS Central Science  (2023)

\bibitem{suppa2022voice}
Suppa, A., Costantini, G., Asci, F., Di~Leo, P., Al-Wardat, M.S., Di~Lazzaro, G., Scalise, S., Pisani, A., Saggio, G.: Voice in parkinson's disease: a machine learning study. Frontiers in Neurology  \textbf{13},  831428 (2022)

\bibitem{goetz2010movement}
Goetz, C.: Movement disorder society-unified parkinson's disease rating scale (mds-updrs): une nouvelle {\'e}chelle pour l’{\'e}valuation de la maladie de parkinson (2010)

\bibitem{Thieu_PerMetrics_A_Framework_2024}
Thieu, N.V.: {PerMetrics: A Framework of Performance Metrics for Machine Learning Models}. Journal of Open Source Software  \textbf{9}(95), ~6143 (Mar 2024). \doi{10.21105/joss.06143}, \url{https://joss.theoj.org/papers/10.21105/joss.06143}

\bibitem{van2023mealpy}
Van~Thieu, N., Mirjalili, S.: Mealpy: An open-source library for latest meta-heuristic algorithms in python. Journal of Systems Architecture  (2023). \doi{10.1016/j.sysarc.2023.102871}

\bibitem{mediumArticle}
Vidhya, A.: Mae, mse, rmse, coefficient of determination, adjusted r-squared: Which metric is better? (2022), \url{medium.com/analytics-vidhya/mae-mse-rmse-coefficient-of-determination-adjusted-r-squared-which-metric-is-better-cd0326a5697e}

\bibitem{vladik2014estimate}
Vladik, K., Nguyen~Hung, T., Rujira, O.: How to estimate forecasting quality: a system-motivated derivation of symmetric mean absolute percentage error (smape) and other similar characteristics.(2014)  (2014)

\bibitem{mediumArticle1}
Vidhya, A.: Mae, mse, rmse, coefficient of determination, adjusted r-squared: Which metric is better? (2022), \url{https://scholarworks.utep.edu/cs_techrep/865}

\bibitem{ahmed2020prediction}
Ahmed, S.: Prediction of Rate of Disease Progression in Parkinson’s Disease Patients based on RNA-Sequence using Deep Learning. Ph.D. thesis, Universit{\'e} d'Ottawa/University of Ottawa (2020)

\bibitem{guadalajarabenchmarking}
Guadalajara, J.: Benchmarking of machine learning models for parkinson’s disease progression prediction

\end{thebibliography}
\end{document}